\def\year{2019}\relax
\documentclass[letterpaper]{article} 
\usepackage{aaai19}
\usepackage{times}  
\usepackage{helvet}  
\usepackage{courier}  
\usepackage{url}  
\usepackage{graphicx}  
\usepackage[ruled,linesnumbered,vlined]{algorithm2e}
\usepackage{array}
\usepackage{amsmath}
\usepackage{adjustbox}
\usepackage{color}
\newcolumntype{?}{!{\vrule width 1pt}}

\newcommand{\rpm}{\raisebox{.2ex}
{$\scriptstyle\pm$}}  

\begin{document}

\title{Combinets: Creativity via Recombination of Neural Networks}
\author{Matthew Guzdial and Mark Riedl\\
School of Interactive Computing\\
Georgia Institute of Technology\\
mguzdial3@gatech.edu and riedl@cc.gatech.edu\\
}
\maketitle

\begin{abstract}
One of the defining characteristics of human creativity is the ability to make conceptual leaps, creating something surprising from typical knowledge.
In comparison, deep neural networks often struggle to handle cases outside of their training data, which is especially problematic for problems with limited training data.
Approaches exist to transfer knowledge from problems with sufficient data to those with insufficient data, but they tend to require additional training or a domain-specific method of transfer.
We present a new approach, conceptual expansion, that serves as a general representation for reusing existing trained models to derive new models without backpropagation.
We evaluate our approach on few-shot variations of two tasks: image classification and image generation, and outperform standard transfer learning approaches.

\end{abstract}

\section{Introduction}

Modern deep learning systems perform well with large amounts of training data on known classes but often struggle otherwise. This is a general issue given the invention or discovery of novel classes, rare or illusive classes, or the imagining of fantastical classes. For example, if a new traffic sign were invented tomorrow it would have a severe, negative impact on autonomous driving efforts until there were enough examples for a learning system to recognize this new class with confidence.

Deep learning success has depended more on the size of datasets than the strength of algorithms \cite{UnreasonableEffectivenessOfData}. 
We observe that a significant amount of training data for many classes exists. 
But there are also many novel, rare, or fantastical classes with insufficient data that can be understood as derivations or combinations of existing classes. 
For example, consider a pegasus, a fantastical creature that appears to be a horse with wings, and therefore can be thought of as a combination of a horse and a bird. 
If we suddenly discovered a pegasus and only had a few pictures, we couldn't train a typical neural network classifier to recognize a pegasus as a new class nor a generative adversarial network to create new pegasus images. 
However we might be able to approximate both models given appropriate models trained on horse and bird data.

Various approaches exist to reuse knowledge from larger datasets for problems with smaller datasets, such as zero-shot and transfer learning. 
In these approaches, knowledge from a source model trained on a large dataset is applied to a target problem by either retraining the network on the target dataset \cite{levy2012teaching} or leveraging sufficiently general or authored features to handle new classes \cite{xian2017zero}. 
The latter of these two approaches is not guaranteed to perform well depending on source and target problems, and the former of these is limited in terms of what final target models can be learned.

{\em Combinational creativity} 
is
the type of creativity humans employ when combining ideas \cite{boden2004creative}. Many algorithms exist that attempt to represent this process, but they have historically required hand-authored graphical representations of input concepts with combination only occurring across symbolic values \cite{fauconnier2001conceptual}. 
A neural network is a large, complex graph of numeric values derived from data.
If combinational creativity techniques can be applied to recombine trained neural networks, they could potentially address the pegasus problem and improve few-shot recognition and generation of new classes without the introduction of outside knowledge or heuristics.

We introduce a novel representation, \textit{conceptual expansion}, that allows for the recombination of an arbitrary number of learned models into a final model without additional training.
In the domains of image recognition and image generation we demonstrate how recombination via conceptual expansion outperforms standard transfer learning approaches for fixed neural network architectures. 
The remainder of the paper is organized as follows: first we discuss related work and differentiate this technique from similar approaches for few-shot problems. Second, we discuss the conceptual expansion representation in detail and the search-based approach we employ to construct them in this paper. Third, we present a variety of experiments to demonstrate the limitations and advantages of the approach. We end with conclusions and future work.

\section{Related Work}

\subsection{Combinational Creativity}

\textit{Combinational creativity} represents a particular set of approaches for knowledge reuse through recombining existing knowledge and concepts for the purposes of inventing novel concepts \cite{boden2004creative}. There have been many approaches to combinational creativity over the years. Case-based reasoning (CBR) represents a general AI problem solving approach that relies on the storage, retrieval, and adaption of existing solutions \cite{de2005retrieval}. The adaption function has lead to a large class of combinational creativity approaches~\cite{wilke1998techniques,fox2009exploring,manzano2011amalgam}. These techniques tend to be domain-dependent, for example for the problem of text generation or tool creation \cite{hervas2006case,sizov2015evidence}. Murdock and Goel \shortcite{murdock2001meta} combine reinforcement learning with case-based reasoning, which aligns with our work to combine combinational creativity and machine learning research. However, the technique does not combine classes.

The area of belief revision, modeling how beliefs change, includes a function to merge prior existing beliefs with new beliefs \cite{cojan2009belief,konieczny2011logic,fox2009exploring}. Amalgams represent an extension of this belief merging process that looks to output the simplest combination \cite{ontanon2010amalgams}. The mathematical notion of convolution has been applied to blend weights between two neural nets in work that parallels our desire to combine combinational creativity and machine learning, but with inconclusive results \cite{thagard2011aha}. 

Conceptual blending is perhaps the most popular computational creativity technique, though it has traditionally been limited to hand-authored input \cite{fauconnier2001conceptual}. Li et al. \shortcite{li2012goal} introduced goals to conceptual blending, which parallels our usage of training data to derive the structure of a combination. However, conceptual blending only deals with symbolic values, which makes it ill-suited to machine-learned models. Visual blending \cite{cunha2017pig}, blends components of images using conceptual blending and parallels are use of combinational creativity with Generative Adversarial Networks, however it requires hand-defined components and combines images instead of models. Guzdial and Riedl \shortcite{guzdial2016learning} utilized conceptual blending to recombine machine-learned models of video game level design by treating all numbers as ordinal values, but their approach does not generalize to neural networks.

Combinational creativity algorithms tend to have many possible valid outputs. This is typically viewed as undesirable, with general heuristics or constraints designed to pick a single correct combination from this set \cite{fauconnier2001conceptual,ontanon2010amalgams}. This limits the potential output of these approaches, we instead employ a domain-specific heuristic criteria to find an optimal combination.

\subsection{Knowledge Reuse in Neural Networks}

A wide range of prior approaches exist for the reuse or transfer of knowledge in neural networks, such as zero-shot, one-shot, and few-shot learning \cite{xian2017zero,fei2006one}, domain adaption \cite{daume2009frustratingly}, and transfer learning \cite{lampert2009learning,wang2016learning}. 
These approaches either require an additional set of features for transfer, or depend upon backpropagation to refine learned features from some source domain to a target domain. In the former case these additional transfer features can be hand-authored \cite{lampert2009learning,kulis2011you,ganin2016domain} or learned \cite{levy2012teaching,norouzi2013zero,mensink2014costa,ba2015predicting,elhoseiny2017link}. 
In the case of requiring additional training these approaches can freeze all weights of a network aside from a final classification layer or can tune all the weights of the network with standard training approaches \cite{wong2016sequence,li2017large}. As an alternative one can author an explicit model of transfer such as metaphors \cite{levy2012teaching} or hypotheses \cite{kuzborskij2013stability}. To the best of our knowledge no work exists that attempts few-shot training of generative adversarial networks (GANs), though some work exists at exploring the space between distributions of classes \cite{cheongcan}.

Kuzborskij et al. \shortcite{kuzborskij2013n} investigate the same n to n+1 multiclass transfer learning problem as our image classification experiments, and make use of a combination of existing trained classifiers. However, their approach makes use of Support Vector Machines with a small feature-set and only allows for linear combinations. Rebuffi et al. \shortcite{rebuffi2017icarl} extended this work to convolutional neural nets, but still requires retraining via backpropagation. Chao et al. \shortcite{chao2016empirical} demonstrated that average visual features can be used for zero-shot learning, which represents a domain independent zero-shot learning measure that does not require human authoring or additional training.

One alternative to reusing learned knowledge in neural networks, is to extend a dataset to new classes using query expansions and the web \cite{divvala2014learning,yao2017exploiting} . However, we are interested in problems in which no additional training data exists, even online, due to the class in question being new, fantastical, or rare, and how existing learned features can be adapted.

\section{Conceptual Expansion}

Imagine tomorrow we discover that a pegasus exists. 
Initially we lack enough images of this newly discovered flying horse to build a traditional classifier or image generator. 
However, suppose we have neural network classifiers and generators trained on classes including horses and birds. 
Conceptual expansion allows us to reuse the learned features from machine learned model(s) to produce new models without additional training or additional transfer features.

The intuition behind conceptual expansion is that it allows us to derive a high-dimensional, parameterized search space from an arbitrary number of pretrained input models, where each point of this search space is a new model that can be understood as a being some degree of combination or variation of the input models. 
Each point of this space---each combined model---is a valid conceptual expansion.
We can consider the case where a class ($c_X$) is a combination of other classes ($c1, ...c_n$) and that the learned features of models of classes $c_1, ...c_n$ can be recombined to create the features of a model of $c_X$. In these cases, we hypothesize that conceptual expansions can represent models one cannot necessarily discover using conventional machine learning techniques with the available data. 
Furthermore, we hypothesize that these conceptual expansion models may perform better on specific tasks than standard models in cases with small amounts of available data, such as identifying or generating new classes of objects.
We can use a heuristic informed by this small amount of training data to guide the search for our final conceptual expansion.
This process is inspired by the human ability to make conceptual leaps, but is not intended as an accurate recreation.

A conceptual expansion of concept $X$ is represented as the following function:
\begin{equation} \label{eq1}
  \mathrm{CE^X}(F,A) = a_1*f_1+a_2*f_2... a_n*f_n
\end{equation}

\noindent
Where $F=\{f_1, ...f_n\}$ is the set of all mapped features and $A=\{a_1, ... a_n\}$ is a filter representing what of and what amount of mapped feature $f_i$ should be represented in the final conceptual expansion. 
In the ideal case $X=CE^X$ (e.g. a combined model of birds and horses equals our ideal pegasus model). 
The exact shape of $a_i$ depends upon the feature representation. If features are symbolic, $a_i$ can have values of either 0 or 1 (including the mapped feature or not), or vary from 0 to 1 if features are numeric or ordinal. Note that for numeric values one may choose a different range (e.g. -1 to 1) dependent on the domain. If features are matrices, as in a neural net, each $a_i$ is also a matrix. In the case of matrices the multiplication is an element-wise multiplication or Hadamard product. As an example, in the case of neural image recognition, $\{f_1, ..., f_n\}$ are the variables in a convolutional neural network learned via backpropagation. Deriving a conceptual expansion is the process of finding an $A$ for known features $F$ such that $\mathrm{CE^X}(\cdot)$ optimizes a given objective or heuristic towards some target concept $X$. 

In this representation, the space of conceptual expansions is a multidimensional, parameterized search space over possible combinations of our input models. 
There exists an infinite number of possible conceptual expansions for non-symbolic features, which makes na\"ively deriving this representation ill-advised. 
Instead, as is typical in combinational creativity approaches, we first derive a {\em mapping}. The mapping determines what particular knowledge---in this case the weights and biases of a neural network---will be combined to address a novel case. 
It will serve as an informed starting point that can then be optimized in the space of possible conceptual expansions.

The mapping is the collection of existing class knowledge we will combine from our knowledge base to represent the novel class knowledge initially, and the degree of inclusion of each class. In the aforementioned pegasus example it is unlikely one would have a trained image recognition model that only recognized horses and birds. For example the widely used CIFAR-10 dataset \cite{krizhevsky2009learning} contains ten classes, including horses and birds. If we were to use the CIFAR-10 dataset as our starting point, some of the information from some of the 10 classes might be useful, but much of it likely isn't. We differentiate between these two cases with the mapping. The mapping allows us to select only the portions of a model that will contribute to the recognition of the new class, which can then be used to determine a starting point for searching the space of conceptual expansions.

Given a mapping, we construct an initial conceptual expansion---a set of $\{f_1, ..., f_n\}$ and an $A=\{a_1, ..., a_n\}$---that is iterated upon to optimize for domain specific notions of quality (in the example pegasus case image recognition accuracy). 
We discuss the creation of the mapping in Section~\ref{ref:mapping} and the refinement of the conceptual expansion in Section~\ref{ref:ce}. 


\subsection{Mapping Construction}
\label{ref:mapping}

Constructing the initial mapping is relatively straightforward.
As input we assume we have an existing trained model or models (CifarNet trained on CIFAR-10 for the purposes of this example \cite{krizhevsky2009learning}), and data for a novel class (whatever pegasus images we have). We construct a mapping with the novel class data by looking at how the model or models in our knowledge base perform on the data for the novel class.
The mapping is constructed according to the ratio of the new images classified into each of the old classes. 
For example, suppose we have a CifarNet trained on CIFAR-10 and we additionally have four pegasus images.
Further suppose that CifarNet classifies two of the four pegasus images as a horse and two as a bird. 
We construct a mapping of: $f_1$ consisting of the weights and baises associated with the horse class, and $f_2$ consisting of the weights and biases associated with the bird class. We initialize the alpha values for both variables to all be 0.5---the classification ratio---meaning a floating point value for the biases and a matrix for the weights. This leads to a final classification for our pegasus images that relies on half of the weights of the house class and half of the weights of the bird class. 


\subsection{Conceptual Expansion Search}
\label{ref:ce}
\IncMargin{1em}
\begin{algorithm}[tb]
\footnotesize
\SetKwData{MaxExpansion}{maxE}\SetKwData{Model}{model}\SetKwData{Mapping}{m}\SetKwData{Score}{score}\SetKwData{TotalData}{data}\SetKwData{MaxScore}{maxScore}\SetKwData{Improving}{improving}\SetKwData{Visited}{v}\SetKwData{Neighbor}{n}\SetKwData{NeighborScore}{s}\SetKwData{OldMax}{oldMax}

\SetKwFunction{GetDefaultExpansion}{DefaultExpansion}\SetKwFunction{GetNeighbor}{GetNeighbor}\SetKwFunction{Heuristic}{Heuristic}\SetKwFunction{Max}{max}
\SetKwInOut{Input}{input}\SetKwInOut{Output}{output}
\Input{available data $data$, an initial model $model$, a mapping $m$, and a score $score$}
\Output{The maximum expansion found according to the heuristic}
\BlankLine

\MaxExpansion$\leftarrow$ \GetDefaultExpansion{\Model}+\Mapping\;
\MaxScore$\leftarrow$ \Score\;
\Visited$\leftarrow$ $[$\MaxExpansion$]$\;
\Improving$\leftarrow$ 0\;

\While{\Improving $<$ 10}{
  \Neighbor$\leftarrow$ \MaxExpansion.\GetNeighbor{\Visited}\;
    \Visited$\leftarrow$ \Visited $+$ \Neighbor\;
    \BlankLine 
    \NeighborScore$\leftarrow$ \Heuristic{\Neighbor, \TotalData}\;
    \OldMax$\leftarrow$ \MaxScore
    \MaxScore, \MaxExpansion$\leftarrow$ \Max{$[$\MaxScore, \MaxExpansion$]$, $[$\NeighborScore, \Neighbor$]$}\;
    \Improving$\leftarrow$\OldMax $<$ \MaxScore?0:\Improving++

}

return \MaxExpansion\;
\caption{Conceptual Expansion Search}\label{algo_conceptualExpansionSearch}
\end{algorithm}\DecMargin{1em}
\normalsize

The space of potential conceptual expansions is massive, and the mapping construction stage gives us an initial starting point in this space from which to search. We present the pseudocode for the Conceptual Expansion Search in Algorithm 1. Line 1 finds a default expansion plus the mapping information. The exact nature of this depends on the final network architecture. For example, the mapping may overwrite the entirety of the network if the input models and final model have the same architecture or just the final classification layer if not (as in the case of adding an additional class). 
This initial conceptual expansion will be a linear combination of the existing knowledge, but the final conceptual expansion need not be a linear combination. 
The default expansion is an expansion equivalent to the original model(s), in that each variable is replaced by an expanded variable with its original $f_i$ value and an $a_i$ of 1.0 (or matrix of 1.0's). This means that the initial expansion is functionally identical to the original model, beyond any weights impacted by the mapping. 

Once we have a mapping we search for a set of $F$ and $A$ for which the conceptual expansion performs well on a domain-specific measure $Heuristic$ (e.g. pegasus classification accuracy). For the purposes of this paper we implement a greedy optimization search that checks a fixed number of neighbors before the search ends. The $GetNeighbor$ function randomly selects between one of the following: altering a single element of a single $a_i$, replacing all of the values of a single $a_i$ replacing values of $x_i$ with a randomly selected alternative $x_j$, or adding an addition $x_i$ and corresponding random $a_i$ to an expanded variable. 
The final output of this process is the maximum scoring conceptual expansion found during the search.
For the purposes of clarity we refer to these conceptual expansions of neural networks as {\em combinets}.

Our initial refinement algorithm is a random search for our initial investigation of conceptual expansions, as our focus for this paper is the representation, not the optimization method.
It is possible that alternative means of searching the space of conceptual expansions may find better conceptual expansions and improve on the baselines we establish in the next section.

\begin{table*}[tbh]
\centering
\caption{A table with the average test accuracy for the first experiment. The orig. column displays the accuracy for the 10,000 test images for the original 10 classes of CIFAR-10. The 11th column displays the accuracy for the CIFAR-100 test images.}
\begin{adjustbox}{width=\linewidth}
\begin{tabular}{|l?c|c?c|c?c|c?c|c?c|c|}
  \hline
    & \multicolumn{2}{|c|}{100} & \multicolumn{2}{|c|}{50} 
    & \multicolumn{2}{|c|}{10}  & \multicolumn{2}{|c|}{5} & \multicolumn{2}{|c|}{1} \\
  \hline
  \textbf{Fox}  & 11th & orig. & 11th & orig. & 11th & orig. & 11th & orig. & 11th & orig. \\
  \hline
  combinet  & \textbf{34.0\rpm3.5} & 81.8\rpm2.2 & \textbf{26.0\rpm5.2} & 81.59\rpm1.9 & \textbf{28.3\rpm3.5} & 79.1\rpm1.6 & \textbf{23.0\rpm8.5} & 80.6\rpm1.2 & \textbf{12.0\rpm9.8} & 80.7\rpm7.2 \\
  \hline
  standard & 7.0\rpm2.7 & 62.04 & 0.0\rpm0.0 & 62.17 & 0.0\rpm0.0 & 62.34 & 0.0\rpm0.0 & 62.44 & 0.0\rpm0.0 & 76.44\rpm3.5 \\
  \hline
  transfer & 5.0\rpm4.3 & \textbf{87.2\rpm0.5} & 0.0\rpm0.0 & \textbf{87.9\rpm0.2} & 0.0\rpm0.0 & \textbf{88.1\rpm0.4} & 0.0\rpm0.0 & \textbf{87.7\rpm0.2} & 0.0\rpm0.0 & \textbf{88.0\rpm1.1} \\
  \hline
    zero-shot  & 11.0\rpm0.7 & 86.2\rpm0.4 & 11.0\rpm1.0 & 86.2\rpm0.8 & 9.6\rpm2.3 & 86.2\rpm0.2 & 10.0\rpm4.6 & 86.0\rpm1.4 & 6.0\rpm3.3 & 83.2\rpm2.5 \\
  \hline
  \hline
  \textbf{Plain}  & 11th & orig. & 11th & orig. & 11th & orig. & 11th & orig. & 11th & orig. \\
  \hline
  combinet  & \textbf{53.0\rpm10.0} & 84.0\rpm3.6 & \textbf{45.7\rpm7.6} & 84.2\rpm7.8 & \textbf{31.3\rpm22.0} & 83.9\rpm2.4 & \textbf{28.3\rpm12.6} & 82.3\rpm2.2 & \textbf{23.0\rpm17.4} & 84.0\rpm2.4 \\
  \hline
  standard  & 50.0\rpm7.7 & 62.54 & 42.0\rpm3.2 & 62.18 & 16.0\rpm12.8 & 61.67 & 0.0\rpm0.0 & 62.27 & 0.0\rpm0.0 & 62.27 \\
  \hline
transfer  & 4.5\rpm3.0 & \textbf{86.92} & 0.0\rpm0.0 & \textbf{86.91} & 0.0\rpm0.0 & \textbf{86.96} & 0.0\rpm0.0 & \textbf{87.20} & 0.0\rpm0.0 & \textbf{87.20} \\
  \hline
 zero-shot  & 23.0\rpm0.7 & 86.2\rpm0.5 & 23.6\rpm1.1 & 86.2\rpm0.3 & 22\rpm2.8 & 86.1\rpm13.9 & 18.6\rpm3.8 & 83.7\rpm3.4 & 15.6\rpm7.3 & 82.7\rpm2.9 \\
  \hline
\end{tabular}
\end{adjustbox}
  \label{tab:foxResults}
\end{table*}

\section{CifarNet Experiments}

In this section we present a series of experiments meant to demonstrate the strengths and limitations of conceptual expansions for image classification with deep neural networks. 
We chose CIFAR-10 and CIFAR-100 \cite{krizhevsky2009learning} as the domain for this approach as these represent well-understood datasets. 
It is not our goal to achieve state of the art on CIFAR-10 or CIFAR-100; we instead use these datasets to construct problems in which a system must identify images of a class not present in some initial training set given limited training data on the novel class. For the deep neural network model we chose CifarNet~\cite{krizhevsky2009learning}, again due to existing understanding of its performance on the more traditional applications of these datasets. 
We intentionally choose not to make use of a larger dataset like ImageNet or a larger architecture \cite{imagenet_cvpr09}, as we aim to compare how our approach constructs final features given a limited set of input features, compared to other approaches that transfer features.
We do not include a full description of CifarNet but note that it is a two-layer convolutional neural net with three fully-connected layers. 

For each experiment, we ran our conceptual expansion search algorithm ten times and took the most successful conceptual expansion found across the ten runs in terms of training accuracy. 
We did this to ensure we had found a near optimal conceptual expansion.
We note that this approach was still many times faster than initially training the CifarNet on CIFAR-10 with backpropagation.

Our first experiment expands a CifarNet trained on CIFAR-10 to recognize one additional class selected from CIFAR-100 that is not in CIFAR-10.
We vary the size of slices of the training data for the newly introduced class,
which allows us to evaluate the performance of recombination via conceptual expansions under a variety of controlled conditions.
Our second experiment fully expands a CifarNet model trained on CIFAR-10 to recognize the one-hundred classes of CIFAR-100 with limited training data. 
Finally, we investigate the running example throughout this paper: expanding a CifarNet model trained on CIFAR-10 to classify pegasus images.

\subsection{CIFAR-10 + Fox/Plain}

For our initial experiment we chose to add fox and plain (as in a grassy field) recognition to the CifarNet, as these classes exist within CIFAR-100, but not within CIFAR-10 
(CIFAR-10 is made up of the classes: airplane, automobile, bird, cat, deer, dog, frog, horse, ship, and truck).
We chose foxes and plains for this initial case study because they represented illustrative examples of conceptual expansion performance. 
There exists a number of classes in CIFAR-10 that can be argued to be similar to foxes, but no classes similar to plains.

For training data we drew from the 50,000 training examples for the ten classes of CIFAR-10, adding a varying number of training instance of fox or plain. 
For test data we made use of the full 10,000 CIFAR-10 test set and the 100 samples in the CIFAR-100 test set for each class. 
For each size slice of training data (i.e. 1, 5, 10, and 100) we constructed five unique random slices. 
We chose five for consistency across all the differently sized slices, given that there was a maximum of 500 training images for fox and plan, and our largest slice size was 100.
We present the average test accuracy across all approaches and with all sample sizes in Table \ref{tab:foxResults}. 
This table shows results when we provide five slices of fox or plain images in the quantities of 1, 5, 10, 50, or 100. 
For each slice, we provide the accuracy on the original CIFAR-10 images and the accuracy of identifying the 11th class (either fox or plains).

We evaluate against three baselines. 
Our first baseline (standard) trains CifarNet with backpropagation with stratified branches on the 10,000 CIFAR-10 images and newly introduced foxes or plains.
This baseline makes the assumption that the new class was part of the same domain as the other classes as in \cite{daume2009frustratingly}. 
For our second baseline we took inspiration from transfer learning and student-teacher models \cite{wong2016sequence,li2017large,furlanello2017born}, and train an initial CifarNet on only the CIFAR-10 data and then retrain the classification layers to predict the eleventh class with the newly available data. 
We note that transfer learning typically involves training on a larger dataset, such as ImageNet, then retraining the final classification layer. 
However, we wished to compare how these different approaches alter the same initial features towards classifying the new class. 
For our third baseline we drew on the zero-shot approach outlined in \cite{chao2016empirical}, using the average activation of the trained CifarNet on the training data to derive feature classification vectors. 
In all cases we trained the model until convergence. 

There exist many other transfer approaches, but other approaches tend to require additional human authoring of transfer methods or features or an additional dataset to draw from. 
We focus on comparing the behavior of these approaches in terms of altering or leveraging learned features, and so making use of these other approaches would only make this less clear.

As can be seen in Table \ref{tab:foxResults}, the combinet consistently outperforms the baselines at recognizing the newly added eleventh class. 
We note that the expected CifarNet test accuracy for CIFAR-10 is 85\%. 
Combinets achieve the best accuracy on the newly added class while only losing a small amount of accuracy on average on the 10 original classes.
The combinet loss in CIFAR-10 accuracy was almost always due to overgeneralizing. 
The transfer approach did slightly better than the expected CIFAR-10 accuracy, but this matches previously reported accuracy improvements from retraining \cite{furlanello2017born}. 

Foxes clearly confused the baselines, leading to no correctly identified test foxes for the standard of transfer baselines for the lowest values. Compared to plains, foxes had significant overlap in terms of features with cats and dogs. With these smaller sizes samples transfer and standard were unable to learn or adapt suitable discriminatory features. Comparatively, the conceptual expansion approach was capable of combining existing features into new features that were more successfully able to discriminate between these classes. The zero-shot approach did not require additional training and instead made use of secondary features to make predictions, which was more consistent, but still not as successful as our approach in classifying the new class.

Note that combinets do not always outperform these other approaches. For example, the standard approach beats out combinets, getting an average of 83\% accuracy with access to all 500 plain training images, while the combinet only achieves an accuracy of roughly 50\%. This suggests that combinets are only suited to problems with low training data.

\begin{table*}[t]
\centering
\caption{Summary of results for the GAN experiments.}
\footnotesize
\begin{adjustbox}{width=1\textwidth}
\begin{tabular}{|l?c|c?c|c?c|c?c|c?c|c|}
  \hline
    & \multicolumn{2}{|c|}{combiGAN} & \multicolumn{2}{|c|}{combi+N} & \multicolumn{2}{|c|}{combi+T}  & \multicolumn{2}{|c|}{Naive} & \multicolumn{2}{|c|}{Transfer} \\
  \hline
  Samples  & I & KL & I & KL & I & KL & I & KL & I & KL \\
  \hline
  500  & 3.83\rpm0.32 & 0.33 & \textbf{4.61\rpm0.22} & \textbf{0.28} & 3.05\rpm0.23 & 0.31 & 2.98\rpm0.25 & 0.33 & 3.38\rpm0.19 & 1.05 \\
  \hline
  100  & 4.23\rpm0.15 & \textbf{0.10} & 4.38\rpm0.37 & 0.29 & \textbf{4.40\rpm0.19} & 0.43 & 1.76\rpm0.04 & 0.33 & 3.26\rpm0.23 & 0.36 \\
  \hline
  50  & \textbf{4.05\rpm0.24} & 0.22 & 4.03\rpm0.35 & \textbf{0.12} & 1.69\rpm0.05 & 2.36 & 1.06\rpm0.00 & 10.8 & 3.97\rpm0.22 & 0.21 \\
  \hline
  10  & 4.67\rpm0.44 & 0.44 & \textbf{4.79\rpm0.28} & 0.13 & 3.06\rpm0.19 & 1.20 & 1.20\rpm0.01 & 10.8 & 4.40\rpm0.19 & \textbf{0.11} \\
  \hline
\end{tabular}
\end{adjustbox}
  \label{tab:ganResults}
\end{table*}

\subsection{Expanding CIFAR-10 to CIFAR-100}

For the prior experiments we added a single eleventh class from CIFAR-100 to a CifarNet trained on CIFAR-10. 
This experiment looks at the problem of expanding a trained CifarNet from classifying the ten classes of the CIFAR-10 dataset to the one-hundred classes of the CIFAR-100 dataset.

For this experiment we limited our training data to ten randomly chosen samples of each CIFAR-100 class. 
We slightly altered our approach to account for the change in task, constructing an initial mapping for each class individually as if we were expanding a CifarNet to just that eleventh class. 
We utilized the same two baselines as with the first experiment, given the same 1,000 image training set. 

We note that one would not typically utilize CifarNet for this task. 
Even given access to all 50,000 training samples of CIFAR-100 a CifarNet trained using backpropagation only achieves around 30\% test accuracy for CIFAR-100. 
We mean to show the relative scale of accuracy before and after conceptual expansion and not an attempt to achieve state of the art on CIFAR-100 with the full dataset. We tested on the 100,000 test samples available for CIFAR-100.

The average test accuracy across all 100 classes are as follows: the combinet achieves 11.13\%, the naive baseline achieves 1.20\%, the transfer baseline achieves 6.43\%, and the zero-shot baseline achieves 4.10\%. We note that our approach is the only one to do better than chance, and significantly outperforms all the baselines. However no approach reaches anywhere near the 30\% accuracy that could be achieved with full training data for this architecture. 

\subsection{Pegasus}

We return to our running example of an image recognition system that can recognize a pegasus. Unfortunately we lack actual images of a pegasus. To approximate this we collected fifteen photo-realistic, open-use pegasus images from Flickr. Using the same combinet as the above two experiments we tested this approach with a 10-5 training/test split and a 5-10 training/test split. For the former we recognized 4 of the 5 pegasus images (80\% accuracy), with 80\% CIFAR-10 accuracy, and for the latter we recognized 5 of the 10 pegasus images (50\% accuracy) with 82\% CIFAR-10 accuracy. 

\section{DCGAN Experiment}

In this section we demonstrate the application of conceptual expansions to generative adversarial networks (GANs). 
Specifically, we demonstrate the ability to use conceptual expansions to find GANs that can generate images of a class without traditional training on images of that class. 
We also demonstrate how our approach can take as input an arbitrary number of initial neural networks, instead of the one network for the classification experiments. 
We make use of the DCGAN \cite{radford2015unsupervised} as the GAN architecture for this experiment, as it has known performance on a number of tasks.  
We make use of the CIFAR-100 dataset from the prior section and in addition use the Caltech-UCSD Birds-200-2011 \cite{WahCUB_200_2011}, the CAT \cite{zhang2008cat}, the Stanford Dogs \cite{KhoslaYaoJayadevaprakashFeiFei_FGVC2011}, FGVC Aircraft \cite{maji13fine-grained}, and the Standford Cars \cite{KrauseStarkDengFei-Fei_3DRR2013} datasets. We make use of these five datasets as they represent five of the ten CIFAR-10 classes, but with significantly more images and images of higher quality. Sharing the majority of classes between experiments allows us to draw comparisons between results.

We trained a DCGAN on each of these datasets till convergence, then used all five of these models as the original knowledge base for combinets. Specifically, we built mappings by testing the proportion of training samples the discriminator of each GAN classified as real. We then built a combinet discriminator for the target class from the discriminators of each GAN. Finally we built a combinet generator from the generators of each GAN, using the combinet discriminators as the heuristic in traditional GAN-fashion for the conceptual expansion search. We nickname these combinet discriminators and generators {\em combiGANs}. As above we made use of the fox images of CIFAR-100 as our novel class training data, varying the number of available images.

We built two baselines: (1) A naive baseline, which involved training the DCGAN on the available fox images in the traditional manner. (2) A transfer baseline, in which we took a DCGAN trained on the Stanford Dogs dataset and retrained it on the fox dataset. We also built two variations of combiGAN: (1) A combiGAN baseline in which we used the discriminator of the na\"ive baseline as the heuristic for the combinet generator (Combi+N). (2) Same as the last, but using the transfer baseline discriminator (Combi+T). 
We further built a baseline trained on the Stanford Dogs, CAT dataset, and Fox images simultaneously as in \cite{cheongcan}, but found that it did not have any improvement over the other baselines thus we omit it to save space. 
We do not include the zero shot approach of the prior section as it is only suitable for classification tasks.

\begin{figure}[tb]
  \centering
  \includegraphics[width=\columnwidth]{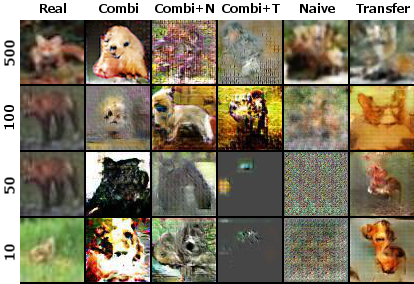}
  \caption{Most fox-like output according to our model for each baseline and sample size.}
  \label{fig:ganFoxes}
\end{figure}

\subsection{CombiGAN Results}

We made use of two metrics: the inception score \cite{salimans2016improved} and Kullback-Leibler (KL) divergence between generated image classification and true image classification distributions. 
We acknowledge that inception score was originally designed for ImageNet;
since we do not train on ImageNet, we cannot use this as an objective score, but we can use it as a comparative metric of objectness. 
For the second metric we desired some way to represent how fox-like the generated images were. 
Thus we made use of the standard classifier trained on 500 foxes, though we could have made use of any classifier in theory. 
We compare the distribution over classes of real CIFAR-100 fox images and the fake images with the KL divergence. 
We generated 10,000 images from each GAN to test each metric. We summarize the results of this experiment in Table \ref{tab:ganResults}.

We note that in almost all cases our approach or one its variations (combi+N and combi+T) outperform the two baselines. In the case with 10 training images the transfer baseline beats our approach on our fox-like measure, but this 0.11 differs only slightly from the 0.13 combi+N value. In Figure \ref{fig:ganFoxes}, we include the most fox-like image in terms of classifier confidence from the training samples (real) and each baseline's output. We note that the combiGAN output had a tendency to retain face-like features, while the transfer baseline tended to revert to fuzzy blobs.

\section{Discussion and Limitations}

Conceptual expansions of neural networks---combinets and combiGANs---outperform standard approaches on problems with limited data without additional knowledge engineering.
We refer to this approach generally as conceptual expansion, which is inspired by the human ability to make conceptual leaps by combining existing knowledge. 
Our contributions in this paper are an initial exploration of conceptual expansion of neural networks; we speculate that more sophisticated optimization search routines than the one provided in this paper may achieve greater improvements.

We anticipate the future performance of conceptual expansions to depend upon the extent to which the existing knowledge base contains relevant information to the new problem and ability for the optimization function to find helpful conceptual expansions. We note that one choice of optimization function could be human intuition, and we have had success hand-designing conceptual expansions for sufficiently small problems. 

Conceptual expansions appear less dependent on training data than existing transfer learning approaches as evidenced by the comparative performance of the approach with low training data, This is further evidenced by those instances where conceptual expansion outperformed itself with less training data. We anticipate further exploration of this in future work. We expect these results to generalize to other domains, but recognize our choice of datasets as a potential limiting factor. CIFAR-10 and CIFAR-100 have very low resolution images (32x32 RGB images). Further, we do not make use of traditional data augmentation techniques such as noising or horizontal flips of the images. We note once again that we chose these datasets for our experiments to focus on feature adaptation.

\section{Conclusions}

We present conceptual expansion, an approach to produce recombined versions of existing machine learned deep neural net models. We ran four experiments of this approach compared to common baselines, and found we were able to achieve greater accuracy with less data. Our technique relies upon a flexible representation of recombination of existing knowledge that allows us to represent new knowledge as a combination of particular knowledge from existing cases. To our knowledge this represents the first attempt at applying a model of combinational creativity to neural networks.

\section{Acknowledgments}

We gratefully acknowledge the NSF for supporting this research under NSF award 1525967.

\bibliographystyle{aaai}
\bibliography{aaai}

\end{document}